\documentclass{article}

\usepackage[preprint]{neurips_2023}

\usepackage[utf8]{inputenc} 
\usepackage[T1]{fontenc}    
\usepackage{hyperref}       
\usepackage{url}            
\usepackage{booktabs}       
\usepackage{amsfonts}       
\usepackage{nicefrac}       
\usepackage{microtype}      
\usepackage{xcolor}         
\usepackage{multirow}
\usepackage{makecell}
\usepackage{colortbl}
\usepackage{booktabs}
\definecolor{lightgray}{rgb}{0.9, 0.9, 0.9}
\definecolor{ballblue}{rgb}{0.13, 0.67, 0.8}

\usepackage{algorithm}
\usepackage{algorithmic}
\usepackage{tablefootnote}
\usepackage{graphicx}
\usepackage{wrapfig}
\usepackage[normalem]{ulem}
\usepackage{bm}
\usepackage{xcolor}
\usepackage{amssymb}
\usepackage{pifont}
\newcommand{\cmark}{\ding{51}}%
\newcommand{\xmark}{\ding{55}}%

\usepackage{ifthen}
\usepackage{xcolor}

\title{Rethinking Compression: Reduced Order Modelling of Latent Features in Large Language Models}

\author{%
  Arnav Chavan\thanks{Equal contribution. Work done while Nahush was an Intern at Nyun AI}~~$^{1,2}$, ~Nahush Lele$^{*1}$, ~Deepak Gupta$^2$  \\
  $^1$Nyun AI  $^2$Transmute AI Lab  \\
  \texttt{arnav.chavan@nyunai.com, guptadeepak2806@gmail.com}\\  
}

\begin{document}

\maketitle

\begin{abstract}


Due to the substantial scale of Large Language Models (LLMs), the direct application of conventional compression methodologies proves impractical. The computational demands associated with even minimal gradient updates present challenges, particularly on consumer-grade hardware. This paper introduces an innovative approach for the parametric and practical compression of LLMs based on reduced order modelling, which entails low-rank decomposition within the feature space and re-parameterization in the weight space. Notably, this compression technique operates in a layer-wise manner, obviating the need for a GPU device and enabling the compression of billion-scale models within stringent constraints of both memory and time. Our method represents a significant advancement in model compression by leveraging matrix decomposition, demonstrating superior efficacy compared to the prevailing state-of-the-art structured pruning method.
\end{abstract}

\section{Introduction}
Recent advances in generative modeling have led to a notable increase in the construction of large language models (LLMs), some of which consist of hundreds of billions of parameters. Despite their commendable accuracy, the associated computational demands are considerable, particularly in terms of GPU memory for inference. In practical applications, there is a growing need to compress these models while minimizing the accompanying performance degradation.

Promising approaches to compress LLMs include pruning \citep{frantar2023sparsegpt, sun2023simple, llmpruner}, quantization \citep{frantar2022gptq,  dettmers2023qlora} and knowledge distillation \citep{wu2023lamini, gu2023knowledge}. Current LLM quantization methods require specific hardware-level support and are not able to reduce MACs and speed up inference time due to expensive quant-dequant operations in LLMs. Knowledge distillation has been shown to perform well in a training-aware fashion on standard deep learning models. 
However, the massive computational resources needed for distillation limits the applicability of such approaches.
Recently, \mbox{\citep{llmpruner}} presented a structured pruning approach designed for LLMs. While this approach is capable of pruning the LLMs with no need for fine-tuning, the drop in performance is significant, and clearly there is a need to explore further in this direction. Moreover, the pruning strategy is not universal and significant effort is needed per neural architecture to identify the prunable structures.

In this paper, we present a novel, practical and training-free approach to model compression which is specifically for large models including LLMs. Referred further as LLM-ROM, our approach performs localized reduced order modelling of the latent features through low-rank decomposition in the feature space and re-parameterization in the weight space. Since LLM-ROM operates layerwise, it does not require any massive model updates and can be executed on small GPU/CPU resources. The simplicity of LLM-ROM facilitates the compression of billion-scale models within stringent constraints of both memory and time. Our early experiments demonstrate that LLM-ROM outperforms existing approaches and can compress LLMs without any fine-tuning. 

\section{Method}

LLM-ROM builds reduced-order model (ROM) layerwise, and for a model with \(L\) layers, the decomposition of the latent feature maps is done in a sequential manner using a calibration data \(\mathbf{X} \in \mathbb{R}^{B \times d_1}\), where $B$ and $d_1$ denote the batch-size and number of input channels, respectively. For the $i^{th}$ layer, denoted as \(L_i\) with weights \(\mathbf{W}_i \in \mathbb{R}^{d_2 \times d_1}\) where \(d_2\) denotes the output channels, we compute the feature map \(\mathbf{Y}_i = \mathbf{W}_i \mathbf{X}_i \in \mathbb{R}^{B \times d_2}\). Following this, the principal components of \(\mathbf{Y}_i\) are then computed through eigenvalue decomposition of the symmetric covariance matrix of \(Y_i\). These components can be represented as \(V_j \in \mathbb{R}^{d_2}\) $\enskip \forall \enskip j \in [1, d_2]$, and the principal component matrix can be represented as \(\mathbf{V} \in \mathbb{R}^{d_2 \times d_2}\), with each row denoting a principal component arranged in the descending order of their eigenvalue.

Depending upon the target rank of the layer, we select only the top \(r\) principal components ranked by their respective eigenvalues. Thus, we index \(V_r = \mathbf{V}[1\rightarrow r, :] \in \mathbb{R}^{r \times d_2}\). Thus, the ROM of this layer can be denoted as \(\mathbf{Y}_i = \mathbf{V}_r^T \mathbf{V}_r \mathbf{W}_i \mathbf{X}_i\). Upon re-parameterization into low-rank matrices, \(\mathbf{W}_{i1} = \mathbf{V}_r^T \in \mathbb{R}^{d_2 \times r}\) and \(\mathbf{W}_{i2} = \mathbf{V}_r \mathbf{W}_i \in \mathbb{R}^{r \times d_1}\), the layer can be decomposed into a sequential combination of two smaller linear layers with weights \(\mathbf{W}_{i1}\) and \(\mathbf{W}_{i2}\) respectively. We consider the ROM of the previous layer to generate inputs for the next layer so that the next layers have prior information of the error introduced in the previous layers for decomposition. Note that the ROM operations are performed on CPU with no requirement for a GPU, and the computational cost associated with it is very small.

\subsection{Layerwise rank computation}
The LLaMA-7B \citep{touvron2023llama} model consists of 32 identical decoder modules (these modules comprise \textgreater 96\% of the total model parameters), each of which consists of seven decomposable weight matrices. Our initial studies showcased that setting a uniform compression budget for all the modules from the very beginning of the model lead to a significant deterioration in model performance; for this reason, we restrict the application of our compression process to a subset of modules. Further, decomposition of the layers of a module introduces errors in the outputs of that layer which get compounded as we move forward in the network; to minimize this, we only compress modules towards the end of the model. Based on these heuristics we perform experiments compressing varying numbers of modules from the end depending on the budget that we need to satisfy for the entire model. 
The specific number of modules to be compressed is empirically determined for each budget. For instance, to achieve an overall budget of 80\%, we conducted experiments compressing only the last 8 modules uniformly with a budget of 0.20, the last 12 modules with a budget of 0.46, and the last 16 modules with a budget of 0.60. Our findings indicated that compressing the last 12 modules yielded the most favorable results for 80\% budget. Similar experiments for 90\% and 50\% budget yield the best results when we compress the last 8 modules with a budget of 0.60 and the last 24 modules with a budget of 0.33 respectively, the results of which are listed in Table \ref{tab:mytable}. Each module is originally composed of 4 weight matrices \(\mathbf{W}_i \in \mathbb{R}^{4096\times 4096}\) from the self attention block and 3 weight matrices \(\mathbf{W}_j \in \mathbb{R}^{4096\times 11008}\) from the feed-forward network (although one of these is transposed, it does not change the computed rank). 
At the end of compression, each weight matrix in the self-attention block is decomposed into two low-rank matrices: \(\mathbf{W}_{i1} \in \mathbb{R}^{4096\times r}\) and \(\mathbf{W}_{i2} \in \mathbb{R}^{r\times 4096}\), where \(r\) takes values of 1228, 954, and 675. Additionally, the weight matrices in the feed-forward network are decomposed into \(\mathbf{W}_{j1} \in \mathbb{R}^{4096\times r}\) and \(\mathbf{W}_{j2} \in \mathbb{R}^{r\times 11008}\) with \(r\) values of 1791, 1373, and 985. These decompositions correspond to three module budget settings of 60\%, 46\%, and 33\%, resulting in overall model budgets of 90\%, 80\%, and 50\%, respectively.

\section{Experiments}
\subsection{Zero-shot performance}
To evaluate the performance of the model in a task-agnostic setting, we employ LLaMA's \citep{touvron2023llama} assessment methodology, performing zero shot task classification across common sense reasoning datasets, including BoolQ \citep{clark-etal-2019-boolq}, PIQA \citep{bisk2020piqa}, HellaSwag \citep{zellers2019hellaswag}, WinoGrande \citep{10.1145/3474381}, ARC-easy \citep{clark2018think}, and ARC-challenge \citep{clark2018think}. We use a batch-size of 512 for the calibration data (Section \ref{calib}) from the training splits of the aforementioned datasets, ensuring no data leakage, and set a maximum sequence length of 128. We set target compression rates of 80\%, and 50\% and compare with LLM-Pruner\footnote{We pick the best performing pruning method from LLM-Pruner \emph{i.e.} Block Pruning.} with and without fine-tuning in Table \ref{tab:mytable}. 

Our LLM-ROM method consistently outperforms LLM-Pruner at 80\% and 50\% compression without any fine-tuning. It is noteworthy that at 80\% budget our method even outperforms fine-tuned LLM-Pruner model, signifying that ROM is able to better extract smaller neural structures and weights from larger counterparts without any gradient updates on the extracted weights. 

\begin{table}[htbp]
    \centering
    \resizebox{\textwidth}{!}{
    \begin{tabular}{lcccccccccc}
        \toprule
        \textbf{Method} & \textbf{Finetune} & \textbf{\#Params} & \textbf{\#MACs} & \textbf{BoolQ} & \textbf{PIQA} & \textbf{HellaSwag} & \textbf{WinoGrande} & \textbf{ARC-e} & \textbf{ARC-c} & \textbf{Average} \\
        \midrule
        LLaMA-7B & - & 6.7B & 423.93G & 76.5 & 79.8 & 76.1 & 70.1 & 72.8 & 47.6 & 70.5 \\ \midrule
        LLM-Pruner & \xmark & 5.4B & 339.60G & 59.4 & 75.6 & 65.3 & 61.3 & 59.2 & 37.1 & 59.7 \\
        LLM-Pruner & \cmark & 5.4B & 339.60G & 69.5 & 76.4 & 68.1 & 65.1 & 63.4 & 37.9 & 63.4 \\
        LLM-ROM & \xmark & 5.4B & 339.99G & 74.5 & 73.8 & 66.6 & 68.1 & 67.2 & 39.8 & 65.0 \\ \midrule
        LLM-Pruner & \xmark & 3.4B & 206.59G & 52.3 & 59.6 & 35.6 & 53.2 & 33.5 & 27.2 & 43.6 \\
        LLM-Pruner & \cmark & 3.4B & 206.59G & 60.3 & 69.3 & 47.1 & 53.4 & 46.0 & 29.2 & 50.9 \\
        LLM-ROM & \xmark & 3.5B & 215.61G & 62.0 & 62.5 & 35.3 & 57.7 & 39.3 & 27.6 & 47.4 \\
        \bottomrule
    \end{tabular}}
    \caption{Comprehensive comparison of our method with LLM-Pruner \cite{} on LLaMA-7B model.}
    \label{tab:mytable}
\end{table}

\subsection{Effect of Batch Size and Sequence Length}
The eigenvalue decomposition of the covariance matrix and the subsequent selection of the principal components require a computation of the outputs of that layer. The batch used to compute this output is a key factor that can influence the generalizability of the layers obtained after decomposition. Principal components computed on a larger sample size will exhibit closer alignment with those of the true distribution. To corroborate this hypothesis, we conducted experiments along two orthogonal directions: one by solely varying the batch-size, and the other with the variation of sequence length, the results for the same are presented in Table \ref{tab:bstable} and \ref{tab:seq_len} respectively. 
\begin{table}[htbp]
    \centering
    \resizebox{\textwidth}{!}{
    \begin{tabular}{lccccccc}
        \toprule
        \textbf{Batch Size}  & \textbf{BoolQ} & \textbf{PIQA} & \textbf{HellaSwag} & \textbf{WinoGrande} & \textbf{ARC-e} & \textbf{ARC-c} & \textbf{Average} \\
        \midrule
       
 512& 74.5& 73.8& 66.6& 68.1& 67.2& 39.8&65.0\\
 128& 72.6& 72.4& 63.2& 66.3& 61.1& 37.7&62.2\\
 32& 70.2& 68.4& 58.7& 67.2& 55.7& 35.7&59.3\\
    \bottomrule
    \end{tabular}}
    \caption{Effect of  batch size on model performance at a sequence length 128.}
    \label{tab:bstable}
\end{table} 
\begin{table}[htbp]
    \centering
    \resizebox{\textwidth}{!}{
    \begin{tabular}{lccccccc}
        \toprule
        \textbf{Seq. Length}  & \textbf{BoolQ} & \textbf{PIQA} & \textbf{HellaSwag} & \textbf{WinoGrande} & \textbf{ARC-e} & \textbf{ARC-c} & \textbf{Average} \\
        \midrule
       
 128& 74.5& 73.8& 66.6& 68.1& 67.2& 39.8&65.0\\
 64& 66.6& 74.4& 65.7& 67.6& 65.4& 40.1&63.3\\
 32& 66.2& 73.4& 65.1& 67.8& 64.6& 39.5&62.7\\
    \bottomrule
    \end{tabular}}
    \caption{Effect of  sequence length on model performance at batch size 512.}
    \label{tab:seq_len}
\end{table}
\\ From Tables \ref{tab:bstable} and \ref{tab:seq_len} it is evident that a larger batch is beneficial and results in significantly better model generalization and at the same time longer sequence length also aids in maintaining model performance post compression. 
\subsection{Choice of calibration dataset}
\label{calib}
Given that the activations of data from the calibration dataset are used to calculate the covariance matrix, which is subsequently utilized for eigendecomposition, it is reasonable to infer that the model's performance is sensitive to the choice of this dataset. This inference is supported by our findings in the conducted studies where we use three different datasets, namely ARC-challenge \citep{clark2018think}, BookCorpus\citep{DBLP:journals/corr/ZhuKZSUTF15} and a combination of all the common sense task prompts i.e each batch contains an equal number of samples from the six common sense reasoning tasks' datasets used for benchmarking namely BoolQ \citep{clark-etal-2019-boolq}, PIQA \citep{bisk2020piqa}, HellaSwag \citep{zellers2019hellaswag}, WinoGrande \citep{10.1145/3474381}, ARC-easy \citep{clark2018think}, and ARC-challenge \citep{clark2018think}
, as our calibration datasets at a budget of \(80\%\) keeping other hyperparameters such as batch-size and sequence length constant. When creating the calibration datasets we choose samples from a data split which is disjoint from the set upon which evaluation is performed, ensuring there is no data leak. The results of these studies are compiled in Table \ref{tab:caldata}.
\begin{table}[htbp]
    \centering
    \resizebox{\textwidth}{!}{
    \begin{tabular}{lccccccc}
        \toprule
        \textbf{Dataset}  & \textbf{BoolQ} & \textbf{PIQA} & \textbf{HellaSwag} & \textbf{WinoGrande} & \textbf{ARC-e} & \textbf{ARC-c} & \textbf{Average} \\
        \midrule
       
 Combination& 74.5& 73.8& 66.6& 68.1& 67.2& 39.8&65.0\\
 ARC-c& 64.6& 72.5& 63.8& 67.0& 67.8& 40.9&62.8\\
 Book Corpus& 63.2& 73.6& 65.6& 67.7& 63.6& 38.0&61.9\\
    \bottomrule
    \end{tabular}}
    \caption{Comparison of model performance with respect to choice of calibration dataset}
    \label{tab:caldata}
\end{table}
\\The results presented above show the influence of the calibration dataset choice on model performance. It is unsurprising that the dataset, which consists of the combination of all common sense tasks used for benchmarking, exhibits the most favorable relative performance.

\section{Computational Cost}
\label{app-cost}
We conduct ROM of LLaMA-7B \citep{touvron2023llama} on a CPU server with 128 GB RAM and 48-core/96-thread processor. Our current implementation loads the complete model at once; however, it is trivial to perform ROM layerwise and hence can be done in under 10 GB of peak RAM given that only inputs and weights of current layer are loaded and processed into the memory. On an average it takes 13 seconds to perform ROM of each layer of LLaMA-7B \citep{touvron2023llama} which has a total of 224 layers. Overall, it takes 15.8 minutes, 21.8 minutes and 28.9 minutes for 90\%, 80\% and 50\% compression rates respectively.

\section{Conclusion}
In this paper we presented a new direction for LLM compression leveraging reduced order modeling of the latent features. Based on the concept of identifying the finite set of most useful latent feature modes, LLM-ROM is capable of compressing LLMs without the need for any fine-tuning. With no requirement of a GPU during the compression process, LLM-ROM can be efficiently run on a simple CPU machine. Moreover, unlike pruning, LLM-ROM is very generic and does not require manual interference for different model architectures. Based on the presented results, we hope to have paved way for a novel approach to design compressed LLMs in a resource-efficient manner.


\bibliography{iclr2023_conference_tinypaper}

\begin{thebibliography}{14}
\providecommand{\natexlab}[1]{#1}
\providecommand{\url}[1]{\texttt{#1}}
\expandafter\ifx\csname urlstyle\endcsname\relax
  \providecommand{\doi}[1]{doi: #1}\else
  \providecommand{\doi}{doi: \begingroup \urlstyle{rm}\Url}\fi

\bibitem[Bisk et~al.(2020)Bisk, Zellers, Gao, Choi, et~al.]{bisk2020piqa}
Yonatan Bisk, Rowan Zellers, Jianfeng Gao, Yejin Choi, et~al.
\newblock Piqa: Reasoning about physical commonsense in natural language, 2020.

\bibitem[Clark et~al.(2019)Clark, Lee, Chang, Kwiatkowski, Collins, and Toutanova]{clark-etal-2019-boolq}
Christopher Clark, Kenton Lee, Ming-Wei Chang, Tom Kwiatkowski, Michael Collins, and Kristina Toutanova.
\newblock {B}ool{Q}: Exploring the surprising difficulty of natural yes/no questions.
\newblock In Jill Burstein, Christy Doran, and Thamar Solorio (eds.), \emph{Proceedings of the 2019 Conference of the North {A}merican Chapter of the Association for Computational Linguistics: Human Language Technologies, Volume 1 (Long and Short Papers)}, pp.\  2924--2936, Minneapolis, Minnesota, June 2019. Association for Computational Linguistics.
\newblock \doi{10.18653/v1/N19-1300}.
\newblock URL \url{https://aclanthology.org/N19-1300}.

\bibitem[Clark et~al.(2018)Clark, Cowhey, Etzioni, Khot, Sabharwal, Schoenick, and Tafjord]{clark2018think}
Peter Clark, Isaac Cowhey, Oren Etzioni, Tushar Khot, Ashish Sabharwal, Carissa Schoenick, and Oyvind Tafjord.
\newblock Think you have solved question answering? try arc, the ai2 reasoning challenge.
\newblock \emph{arXiv preprint arXiv:1803.05457}, 2018.

\bibitem[Dettmers et~al.(2023)Dettmers, Pagnoni, Holtzman, and Zettlemoyer]{dettmers2023qlora}
Tim Dettmers, Artidoro Pagnoni, Ari Holtzman, and Luke Zettlemoyer.
\newblock Qlora: Efficient finetuning of quantized llms.
\newblock \emph{arXiv preprint arXiv:2305.14314}, 2023.

\bibitem[Frantar \& Alistarh(2023)Frantar and Alistarh]{frantar2023sparsegpt}
Elias Frantar and Dan Alistarh.
\newblock Sparsegpt: Massive language models can be accurately pruned in one-shot.
\newblock In \emph{International Conference on Machine Learning}, pp.\  10323--10337. PMLR, 2023.

\bibitem[Frantar et~al.(2022)Frantar, Ashkboos, Hoefler, and Alistarh]{frantar2022gptq}
Elias Frantar, Saleh Ashkboos, Torsten Hoefler, and Dan Alistarh.
\newblock Gptq: Accurate post-training quantization for generative pre-trained transformers.
\newblock \emph{arXiv preprint arXiv:2210.17323}, 2022.

\bibitem[Gu et~al.(2023)Gu, Dong, Wei, and Huang]{gu2023knowledge}
Yuxian Gu, Li~Dong, Furu Wei, and Minlie Huang.
\newblock Knowledge distillation of large language models.
\newblock \emph{arXiv preprint arXiv:2306.08543}, 2023.

\bibitem[Ma et~al.(2023)Ma, Fang, and Wang]{llmpruner}
X.~Ma, G.~Fang, and X.~Wang.
\newblock Llm-pruner: On the structural pruning of large language models.
\newblock \emph{NeurIPS}, 2023.

\bibitem[Sakaguchi et~al.(2021)Sakaguchi, Bras, Bhagavatula, and Choi]{10.1145/3474381}
Keisuke Sakaguchi, Ronan~Le Bras, Chandra Bhagavatula, and Yejin Choi.
\newblock Winogrande: An adversarial winograd schema challenge at scale.
\newblock \emph{Commun. ACM}, 64\penalty0 (9):\penalty0 99–106, aug 2021.
\newblock ISSN 0001-0782.
\newblock \doi{10.1145/3474381}.
\newblock URL \url{https://doi.org/10.1145/3474381}.

\bibitem[Sun et~al.(2023)Sun, Liu, Bair, and Kolter]{sun2023simple}
Mingjie Sun, Zhuang Liu, Anna Bair, and J~Zico Kolter.
\newblock A simple and effective pruning approach for large language models.
\newblock \emph{arXiv preprint arXiv:2306.11695}, 2023.

\bibitem[Touvron et~al.(2023)Touvron, Lavril, Izacard, Martinet, Lachaux, Lacroix, Rozière, Goyal, Hambro, Azhar, Rodriguez, Joulin, Grave, and Lample]{touvron2023llama}
Hugo Touvron, Thibaut Lavril, Gautier Izacard, Xavier Martinet, Marie-Anne Lachaux, Timothée Lacroix, Baptiste Rozière, Naman Goyal, Eric Hambro, Faisal Azhar, Aurelien Rodriguez, Armand Joulin, Edouard Grave, and Guillaume Lample.
\newblock Llama: Open and efficient foundation language models, 2023.

\bibitem[Wu et~al.(2023)Wu, Waheed, Zhang, Abdul-Mageed, and Aji]{wu2023lamini}
Minghao Wu, Abdul Waheed, Chiyu Zhang, Muhammad Abdul-Mageed, and Alham~Fikri Aji.
\newblock Lamini-lm: A diverse herd of distilled models from large-scale instructions.
\newblock \emph{arXiv preprint arXiv:2304.14402}, 2023.

\bibitem[Zellers et~al.(2019)Zellers, Holtzman, Bisk, Farhadi, and Choi]{zellers2019hellaswag}
Rowan Zellers, Ari Holtzman, Yonatan Bisk, Ali Farhadi, and Yejin Choi.
\newblock Hellaswag: Can a machine really finish your sentence?
\newblock In \emph{Proceedings of the 57th Annual Meeting of the Association for Computational Linguistics}, pp.\  4791--4800, 2019.

\bibitem[Zhu et~al.(2015)Zhu, Kiros, Zemel, Salakhutdinov, Urtasun, Torralba, and Fidler]{DBLP:journals/corr/ZhuKZSUTF15}
Yukun Zhu, Ryan Kiros, Richard~S. Zemel, Ruslan Salakhutdinov, Raquel Urtasun, Antonio Torralba, and Sanja Fidler.
\newblock Aligning books and movies: Towards story-like visual explanations by watching movies and reading books.
\newblock \emph{CoRR}, abs/1506.06724, 2015.
\newblock URL \url{http://arxiv.org/abs/1506.06724}.

\end{thebibliography}
\bibliographystyle{iclr2023_conference_tinypaper}

\end{document}